\documentclass[conf]{new-aiaa}
\usepackage[utf8]{inputenc}

\usepackage{graphicx}
\usepackage{amsmath}
\usepackage[version=4]{mhchem}
\usepackage{siunitx}
\usepackage{longtable,tabularx}
\usepackage{caption}  
\title{A Machine Learning Framework for Fault Detection, Isolation, and Severity Prediction of Autonomous VTOL Aircraft}

\author{
\begin{tabular}{cccc}
\normalfont\textbf{Ripon C. Sarker} & \normalfont\textbf{Pedram H. Dabaghian} & \normalfont\textbf{Raman Goyal} & \normalfont\textbf{Atanu Halder} \\
\normalfont Graduate Research Assistant & \normalfont Graduate Research Assistant & \normalfont Research Scientist & \normalfont Assistant Professor \\
\normalfont Oklahoma State University & \normalfont Oklahoma State University & \normalfont SRI International & \normalfont Oklahoma State University \\
\normalfont Stillwater, OK, USA & \normalfont Stillwater, OK, USA & \normalfont Palo Alto, CA, USA & \normalfont Stillwater, OK, USA \\
\end{tabular}
}

\begin{document}
\twocolumn[
\maketitle

\section*{ABSTRACT}
\noindent
Fault detection in autonomous VTOL aircraft is critical because even minor component degradations can rapidly destabilize multirotor vehicles operating in complex, safety-critical environments, motivating robust fault detection and estimation strategies capable of identifying early signs of rotor damage; however, real-flight fault detection remains challenging due to sensor noise, environmental disturbances, and the nonlinear aerodynamics of multirotor platforms. This study proposes a comprehensive machine-learning framework for rotor fault detection, isolation, and severity prediction using real flight data. A convolutional neural network (CNN) architecture is developed to learn spatio-temporal patterns from multivariate flight dynamics, enabling direct inference of both the faulted rotor and its damage level. The framework is first validated using simulated data generated by a data-generative model, and experimental validation is then performed on a hexacopter by introducing controlled blade-tip breakage. The trained model achieves rotor-wise fault classification accuracies above 99\% and severity estimation accuracy of $\sim$96\% within a $\pm1\%$ tolerance in experimental data, demonstrating strong generalization and supporting real-time health monitoring for autonomous VTOL systems.
\vspace{0.4cm} 
]

\begin{center}
\section*{INTRODUCTION}
\end{center}

In recent years, unmanned aerial vehicles (UAVs) have emerged as a transformative technology with widespread applications in various industries, such as precision agriculture, search and rescue, defense surveillance, including public safety-critical missions such as package delivery, inspection, emergency response, and mapping in Ref. \cite{guo2020precision, Mazzia2020UAVVegetationIndex, Sarker2026EmergencyEVTOL, Puri2005TrafficUAVSurvey, coleman2021development,halder2022understanding, Ahmed2025CoaxialPropellers, Ahmed2026AIAA2480}. In such operations, drones frequently navigate over residential areas, roadways, and commercial spaces. In these environments, even a minor in-flight anomaly can create significant safety risks.  For instance, faults in propellers, motors, or sensors can alter thrust balance, resulting in flight instability, unintended drift, or loss of control, which can ultimately lead to a crash or complete loss of the vehicle. As multirotors rely on tightly coupled control loops, even minor faults can escalate quickly. Therefore, fault detection and isolation (FDI) is essential to ensure safe operation, prevent crashes, and enable fault-tolerant control responses. 

\footnotetext{
Presented at the Vertical Flight Society's Vertical Lift Aircraft Design \& 
Aeromechanics Specialist's Conference, San Jose, CA, USA, Jan. 27--29, 2026. 
Copyright © 2026 by the Vertical Flight Society. All rights reserved.
}

Several studies in the existing literature have investigated fault detection, diagnosis, and mitigation strategies for unmanned aerial vehicles (UAVs), with particular emphasis on identifying faults in specific components such as propeller, sensors, and actuators in Ref.\cite{Guo2018PMI, Zhao2021NeuralFDI, Panitsrisit2011ElevatorFDIA, Puchalski2024PADRE} For instance, Yousefi et al. (Ref. \cite{Yousefi2018DataDriven}) proposed a data-driven fault detection framework for UAV actuator faults using supervised learning on actuator voltage and current signals. Healthy and faulty datasets were collected and Logistic Regression (LR) and Linear Discriminant Analysis (LDA) classifiers were trained to identify actuator health states. Both models achieved comparable classification accuracy (79.2\% for LDA and 78.6\% for LR), demonstrating effective actuator-level monitoring without explicit system modeling. In another case, Guo et al. (Ref. \cite{Guo2018Hybrid}) proposed a hybrid framework for UAV sensor fault diagnosis that combines model-based residual generation using an Extended Kalman Filter (EKF) with data-driven classification using a CNN and achieved up to 99.6\% diagnostic accuracy in simulation, though further validation under real flight conditions is needed. In addition, Ozkat (Ref. \cite{Ozkat2024Vibration}) proposed a vibration-based anomaly detection framework for UAV propeller faults using an unsupervised deep learning approach that combines wavelet scattering with an LSTM autoencoder. Experiments on a hexacopter demonstrated high anomaly detection accuracy without labeled data, though validation was limited to a controlled laboratory setting. In another study, Al-Haddad et al.(Ref. ~\cite{AlHaddad2024Propeller}) developed a deep learning-based propeller fault diagnosis framework using complexity/energy features ranked using a Chi-square ($\chi^2$) test and the Taguchi method. A DNN trained on the selected features achieved up to \textbf{99.6\%} accuracy. Despite strong performance, reliance on handcrafted features and curated datasets may limit scalability to real-time in-flight UAV monitoring. Although these methods represent significant progress in UAV fault detection for specific components—such as sensors, propellers, and motors—they still exhibit significant limitations in terms of generalization and diagnostic depth. Most existing studies focus primarily on binary fault classification (healthy vs. faulty), without quantifying the severity, progression, or uncertainty associated with each fault condition. Few studies considered only the simulation data or controlled environments that limit their practicality.

Some efforts have aimed at developing more integrated fault detection frameworks that monitor multiple UAV components/faults simultaneously rather than focusing on isolated components/faults(Ref. \cite{Lieret2020Redundant,Andrioaia2024BLDC}. For example, Andrioaia and Gaitan (Ref. \cite{Andrioaia2024BLDC} proposed a data-driven machine learning framework for detecting and classifying UAV BLDC motor faults using multi-sensor signals (temperature, current, voltage, and vibration) collected from a custom test stand. KNN, SVM, and Bayesian Network models were evaluated, with SVM achieving the highest accuracy (96\%), followed by KNN (95\%) and BN (91\%), though the results were based on limited laboratory fault configurations. In another study, Lieret et al. (Ref.\cite{Lieret2020Redundant}) proposed a fault detection and fault-tolerant control framework for autonomous multirotor UAVs based on a triple modular redundant flight control architecture with three independent controllers running different hardware/software configurations. Experimental results showed effective detection of malfunctions and rotor imbalances while maintaining stable flight through automatic controller switching, although the redundant setup increases system complexity and may produce false alarms during transitional flight states.Although these studies represent significant progress toward more integrated UAV fault detection frameworks—combining sensor, propeller, and motor-level monitoring—they still fall short in effectively addressing uncertainty modeling and fault severity estimation, both of which are essential for reliable diagnosis under real-world operating conditions.

Another group of studies has focused on addressing the impact of uncertainty in UAV fault detection, recognizing that sensor noise, environmental variability, and modeling errors can significantly affect the diagnostic accuracy in these references \cite{Pan2020MultivariateRegressionUAV, Datta2021StatisticalResidualFDI, Chen2016RobustBackstepping, Iliopoulos2024InflightVibrationFDI}. For example, Sadhu et al.(Ref. ~\cite{Sadhu2020Onboard}) proposed an on-board two-stage deep-learning pipeline for resource-constrained UAVs, where a CNN–BiLSTM autoencoder first detects anomalies from raw IMU streams and then a CNN–LSTM classifier identifies the fault/attack type. The framework was validated using Crazyflie 2.0 flight tests and Microsoft AirSim simulations, achieving $>$90\% anomaly-detection accuracy and 85--88\% experimental fault classification accuracy (93--99\% in simulation), although severity estimation was not addressed. Furthermore, Chen et al.(Ref. ~\cite{Chen2024MotorFaults}) proposed a current-signal-driven small-sample UAV motor fault diagnosis framework using a Broad Learning System combined with a 1-D CNN (BLS-CNN) to address limited labeled fault data. Using a quadrotor current-sensing test rig with nine operating states (healthy plus bearing/rotor/stator faults), BLS-CNN achieves around 90\% accuracy with only a few hundred training windows, though validation was limited to bench-top (unloaded) experiments. In addition, Tong et al.(Ref. ~\cite{Tong2023HybridData}) proposed a hybrid data-driven framework for UAV propeller fault detection that combines simulation and experimental data to mitigate the lack of real fault datasets. An LSTM-based damaged-propeller model was integrated with quadrotor dynamics to generate hybrid training data, and a CNN classifier was trained to locate the faulted rotor. The method achieved $>$99\% accuracy in simulation and 76.3\% on real indoor flight tests; however, the approach is limited in case of predicting the severity of the fault, also adaptation to outdoor flight and uncertain aerodynamic conditions remains necessary for robust deployment. Also,  Zou et al.(Ref. ~\cite{Zou2025PropellerFault}) proposed an IMU-based propeller fault detection method for multirotor UAVs using only acceleration signals, avoiding additional sensors. Simulation and F450 quadrotor experiments showed that z-axis acceleration provided the most fault information (similarity value of \textbf{76.2\%}), although validation was limited to hovering flight. In another study, D’Amato et al.(Ref. ~\cite{DAmato2021Particle}) proposed a particle filtering (PF)-based fault detection and isolation (FDI) framework for UAV IMUs using a duplex setup with two IMUs and an onboard Raspberry Pi, avoiding triplex hardware redundancy. Although lightweight and effective, the method still does not address fault severity estimation, limiting adaptability in highly dynamic conditions.Despite these study's strengths in modeling sensor drift, dynamic disturbances, and performing inference under noisy environments, none of these methods explicitly address fault severity classification, which is essential for distinguishing between incipient, moderate, and critical faults.

Recent studies have moved beyond simple fault detection to address the quantification of fault severity in UAV subsystems. For instance, Avram et al.(Ref. ~\cite{Avram2017Quadrotor}) proposed a nonlinear adaptive estimation-based FDIA framework for quadrotor actuator faults, combining analytical redundancy with adaptive thresholding to detect, isolate, and estimate rotor loss-of-effectiveness (LOE). Indoor experiments using a Qbrain controller and Vicon tracking showed that 10--20\% LOE faults were detected and successfully accommodated to maintain trajectory tracking. However, faults below 6\% LOE were difficult to detect, adaptive gains required manual tuning, and validation was limited to controlled indoor conditions without real-world disturbances.In another case, Pourpanah et al.(Ref.~\cite{Pourpanah2018Anomaly}) proposed a condition monitoring framework for UAV motors and propellers by fusing motor current signature analysis (MCSA) and vibration signal analysis (VSA) with neuro-fuzzy and reinforcement learning models. A Fuzzy ART network detected motor faults with \textbf{95.35\% accuracy}, while a Q-learning-based Fuzzy ARTMAP with Genetic Algorithm (QFAM-GA) performed feature selection and classified propeller damage severity (5\%, 10\%, 15\% breakage). Although effective for both fault detection and severity estimation, the approach was validated only in controlled indoor experiments.In another study, Pose et al.(Ref.~\cite{Pose2024Propeller}) proposed a data-driven framework for detecting, classifying, localizing, and quantifying multirotor propeller damage using only IMU and control command data, without additional sensors or aerodynamic models. Frequency-domain features were processed in a hierarchical scheme using SVMs and neural networks to estimate fault type, rotor location, and severity. Experiments on a DJI F450 (5--40~mm cuts) achieved $>$98\% classification accuracy and typically $<$10\% error in damage magnitude estimation, though validation was performed mainly under controlled indoor conditions.

None of the existing studies simultaneously address UAV fault detection, isolation, and severity prediction, while also validating performance using real-time flight data under environmental disturbances on a hexacopter platform. To bridge this gap, we propose a machine learning framework for fault detection, isolation, and severity prediction tailored to autonomous VTOL aircraft. The proposed framework is designed to detect and isolate rotor faults and estimate fault severity using real flight data collected from a hexacopter operating in the presence of environmental uncertainties. This work establishes a foundation for early fault detection and health management systems in  autonomous VTOL UAVs.

\section{ML-based FDI Model Architecture}
\begin{figure}
    \centering
    \includegraphics[width=1\linewidth]{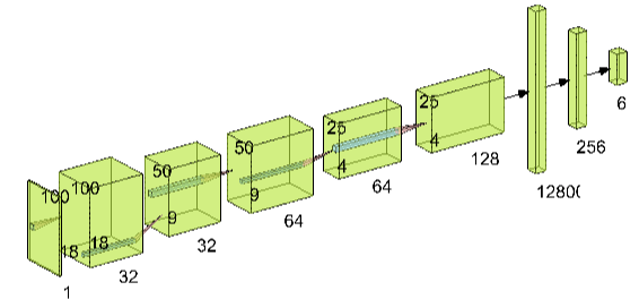}
    \caption{Architecture of CNN regressor model}
    \label{fig:architecture}
\end{figure}

The core of ML-based FDI model is a Convolutional Neural Network (CNN) model that is responsible for fault detection, isolation, and severity evaluation. Among these tasks, the fault detection module plays a critical role, as all subsequent operations—such as fault localization and assessment—depend on the accurate identification of a fault. Figure \ref{fig:architecture} presents the architecture of the convolutional neural network (CNN) regressor developed to detect blade breakage in the rotors of the modeled hexacopter. In addition to determining whether a fault has occurred, the regressor is also capable of identifying the specific rotor on which the fault has taken place.
As shown, the input to the model consists of samples with dimensions $100\times18$, representing time-series data comprising 12 kinematic features of the flight. These features include inertial displacement, Euler angles, body translational and rotational velocities of the hexacopter in three Cartesian directions. The additional six existing features are the rotational speed of the propellers. A more detailed explanation of these features is provided in the following section. The two-dimensional input structure enables the model to capture both spatial and temporal characteristics of the flight dynamics, and consequently, the model's output identifies the specific propeller(s) exhibiting faulty behavior.

The architecture begins with a 2D convolutional layer comprising \(32\) filters of size \(3 \times 3\), followed by a ReLU activation function. To reduce overfitting, $L_2$ regularization with a coefficient of \(0.001\) is applied. A batch normalization layer follows to stabilize and accelerate the training process. Then, max pooling is performed with a pool size of \(2 \times 1\), which down-samples the temporal dimension while preserving the feature dimension.
As it is illustrated in Figure \ref{fig:architecture}, this convolutional structure is repeated two more times, with increasing filter depths of \(64\) and \(128\) respectively in each convolutional layer. Each of these layers is followed by batch normalization and max pooling. The gradual increase in the number of filters allows the network to learn hierarchical representations of the input features, from simple temporal edges to complex feature interactions across multiple time steps.
Following the convolutional blocks, the resulting feature maps are flattened into a one-dimensional vector and passed through a dense layer with \(256\) neurons and ReLU activation.A dropout layer with a relatively high dropout rate of $0.3$ is applied at this stage to enhance generalization by mitigating the co-adaptation of neurons, wherein certain neurons become overly dependent on the activations of others.

The final layer of the network is a fully connected softmax layer with six output neurons, corresponding to the six rotors. Each output neuron represents the probability that the corresponding rotor is faulty. This probability is interpreted as the blade breakage percentage for that rotor.
The model is trained to minimize the sparse categorical cross-entropy loss function. The entire network is optimized using the Adam optimizer with a learning rate of $1\times10^{-6}$.
Classification accuracy is used as the primary performance metric to evaluate the model during both training and validation.

\section{Training and Validation with Simulated Data}
In this section, the training and validation process of the developed CNN fault regressor model is presented. All the mentioned phased are performed using simulation dataset that is obtained by a flight dynamic framework designed for a hexacopter.

\begin{figure}[h!]
\centering
\begin{minipage}{0.48\textwidth}
A comprehensive flight dynamics framework—comprising essential components such as the aerodynamic, flight dynamics, and control modules—was utilized to simulate the flight of a hexacopter executing a mission profile that included a 2-meter ascent, 5-meter forward translation, and 2-meter descent. The resulting simulation data was then used to train, test, and validate the proposed FDI framework. The mission was simulated under various flight scenarios, with each of the six rotors in random fault conditions. Fault was modeled as the breakage on the blade of the rotors and was applied up to $10\%$ of the blade radius.
    
\end{minipage}\hfill
\begin{minipage}{0.5\textwidth}
    \centering
    \includegraphics[width=\linewidth]{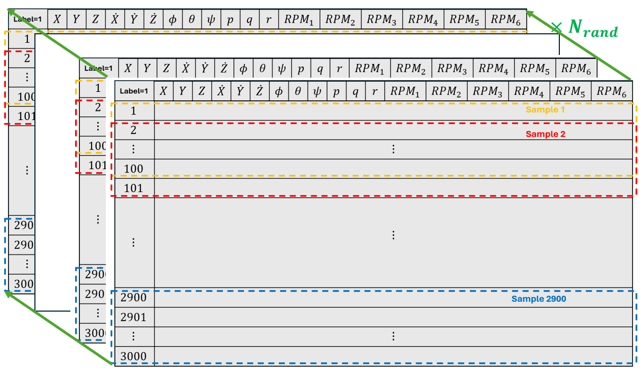}
    \caption{Data structure schematic}
    \label{fig:data}
\end{minipage}
\end{figure}

The structure of the generated dataset is illustrated in Figure~\ref{fig:data}. As shown, each sample comprises 18 features: the inertial position ($x,y,z$), the inertial velocities ($\dot{X}, \dot{Y},\dot{Z}$), the Euler angles $(\phi, \theta, \psi)$, the body rotational velocities ($p,q,r$) , and the rotational speeds of the six propellers \( (\mathrm{RPM}_1, \mathrm{RPM}_2, \mathrm{RPM}_3, \mathrm{RPM}_4, \mathrm{RPM}_5, \mathrm{RPM}_6) \). For each flight scenario, data were recorded over a \( 1 \)~second window with a simulation time step of \( 0.01 \)~seconds, resulting in input samples of size \( 100 \times 18 \).
To augment the dataset, a sliding window approach was employed, generating a new sample at each of the \( 2900 \) possible \( 1 \)-second intervals within the \( 30 \)-second simulation. There were also 200 ($N_{rand}$) random scenario created using different breakage combinations over the six rotors.
Consequently, for each of the \( 200 \) scenarios, \( 2901 \) unique time-history samples were generated, resulting in a total of \( 200 \times 2901 = 580{,}200 \) labeled samples in the dataset. Of these, \( 80\% \) were used for training and testing while the remaining \( 20\% \) were reserved for validation.
Table~\ref{tab:training_params} summarizes the key hyperparameters and dataset characteristics used for training and evaluating the proposed CNN-based FDI model. 
As it is shown, a total of 464,160 samples were used for training, while 116,040 samples were reserved for testing, ensuring a sufficiently large dataset to capture a wide range of fault scenarios and flight conditions. The network was trained for 1000 epochs using a mini-batch size of 64 samples per gradient update. Also, the Adam optimizer was employed with a learning rate of $1\times10^{-5}$ to ensure stable convergence during training. The loss function was selected as the mean squared error (MSE), reflecting the model’s objective of accurately estimating fault severity in addition to fault localization. These training settings were chosen based on extensive empirical testing to balance convergence speed, numerical stability, and generalization performance. Besides, a dropout ratio of $0.3$ was used to improve the overall generalizability of the trained model. Table \ref{tab:training_params} summarizes the hyperparameters used in training process. 

\vspace{10pt}
\begin{table}[ht]
\centering
\resizebox{1\columnwidth}{!}{
\begin{tabular}{|l|c|}
\hline
Parameter & Value \\
\hline
Number of training samples & 464{,}160 \\
Number of testing samples  & 116{,}040 \\
Epochs                     & 1000 \\
Learning rate              & $1 \times 10^{-5}$ \\
Loss function              & Mean Squared Error (MSE) \\
Optimizer                  & Adam (stochastic gradient descent with momentum) \\
Batch size                 & 64 samples per gradient update \\
\hline
\end{tabular}
}
\caption{Training and optimization parameters of the CNN fault regressor model}
\label{tab:training_params}
\end{table}

\begin{figure}[h!]
To evaluate the accuracy of the developed CNN fault regressor model, the predicted breakage percentage values of the rotors were compared to their corresponding true values used in the simulation. Therefore, the prediction error for each rotor is defined as the absolute deviation between the estimated and true breakage levels, and is computed as:
\begin{equation}
\text{Damage error} = 100 \Big| y^{\text{pred}}_{i,r} - y^{\text{true}}_{i,r} \Big|.
\label{eq:prediction_error}
\end{equation}
where $i \in \{1,2,\dots,N_s\}$, $r \in \{1,2,\dots,6\}$, and $N_s$ is the number of samples.
Figure~\ref{fig:test} shows the rotor-wise absolute prediction error for all samples in the test set, which were unseen during training. As evident from the figure, the prediction errors remain consistently low across all six rotors, with the majority of samples exhibiting errors below approximately $1\%$. Occasional higher-error spikes are observed; however, these are sparse and remain bounded within a narrow range, indicating stable and well-controlled prediction behavior. The similarity in error distributions across rotors further suggests that the model’s performance is uniform and does not exhibit rotor-dependent bias.

\begin{minipage}{0.5\textwidth}
    \centering
    \includegraphics[width=\linewidth]{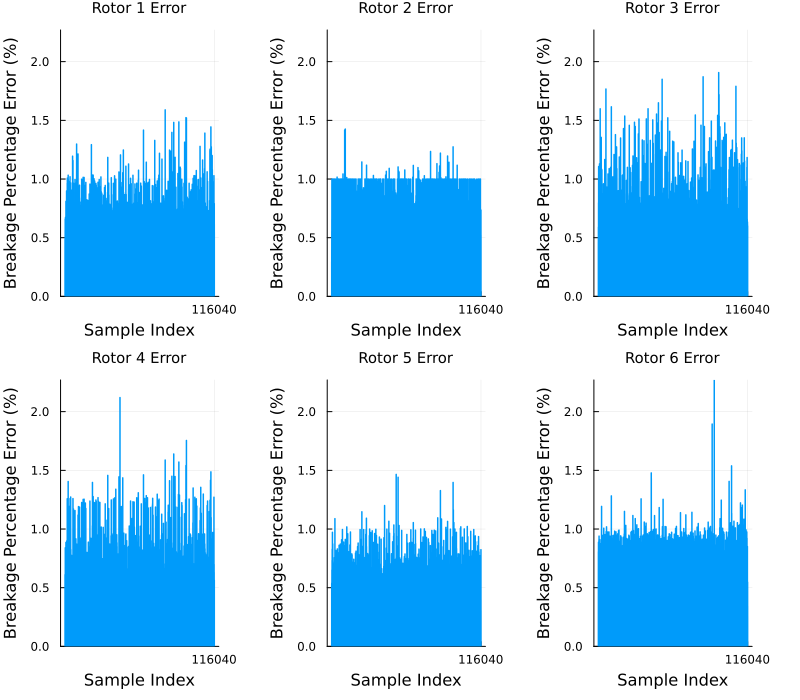}
    \caption{Evaluation of accuracy of CNN model trained with simulated dataset on test dataset}
    \label{fig:test}
\end{minipage}
\end{figure}

\section{Training and Validation with Experimental Data}
\subsection{Experimental Test Platform}
Experimental validation of the proposed machine-learning-based fault detection, isolation, and severity (FDIS) framework was conducted using a custom-built multirotor VTOL testbed configured in a hexacopter layout. The airframe is based on the Flame Wheel 550 (F550) platform, selected for its structural robustness, modular design, and widespread use in experimental multirotor platform. The hexacopter configuration provides actuator redundancy, enabling safe execution of controlled fault scenarios while maintaining flight stability. The propulsion system consists of six identical brushless DC motors, each paired with a matched electronic speed controller (ESC) and fixed-pitch propeller (APC 10x3.8SF). This homogeneous actuator configuration ensures that deviations in system response can be attributed primarily to injected faults rather than geometric or manufacturing asymmetries. The vehicle is powered by a lithium-polymer battery sized to support hover and low-speed maneuvering during repeated experimental trials.

\subsection{Sensor and Data Acquisition System}

\begin{figure}[!h]
    \centering
    \includegraphics[width=0.95\linewidth, height=6cm]{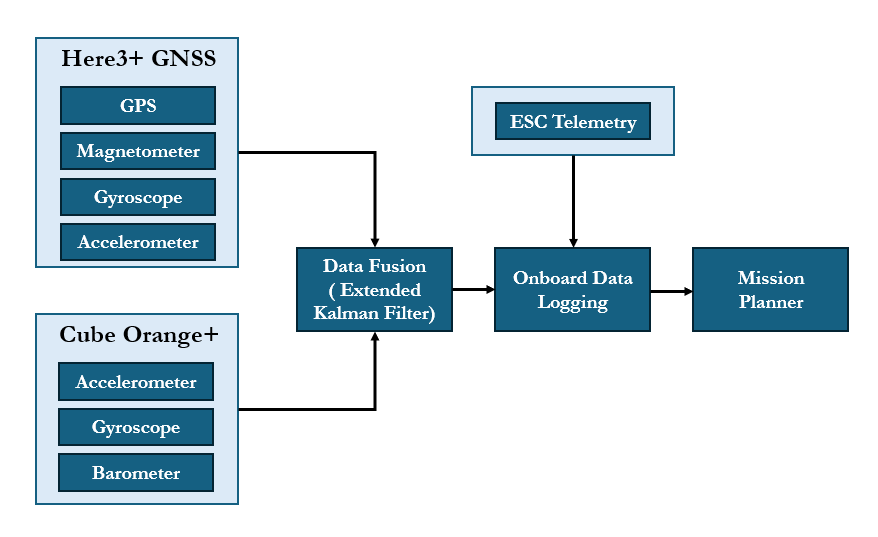}
    \caption{Architecture of the sensor and data acquisition system.  }
    \label{fig:data_acquisition}
\end{figure}

The onboard sensing architecture of the experimental hexacopter is designed to capture the complete state and health information of the vehicle during flight. It integrates both propulsion-level telemetry and high-fidelity navigation data to ensure accurate recording of the vehicle’s dynamic response under different fault conditions.  \begin{minipage}{\linewidth}
    \centering
    \includegraphics[width=\linewidth]{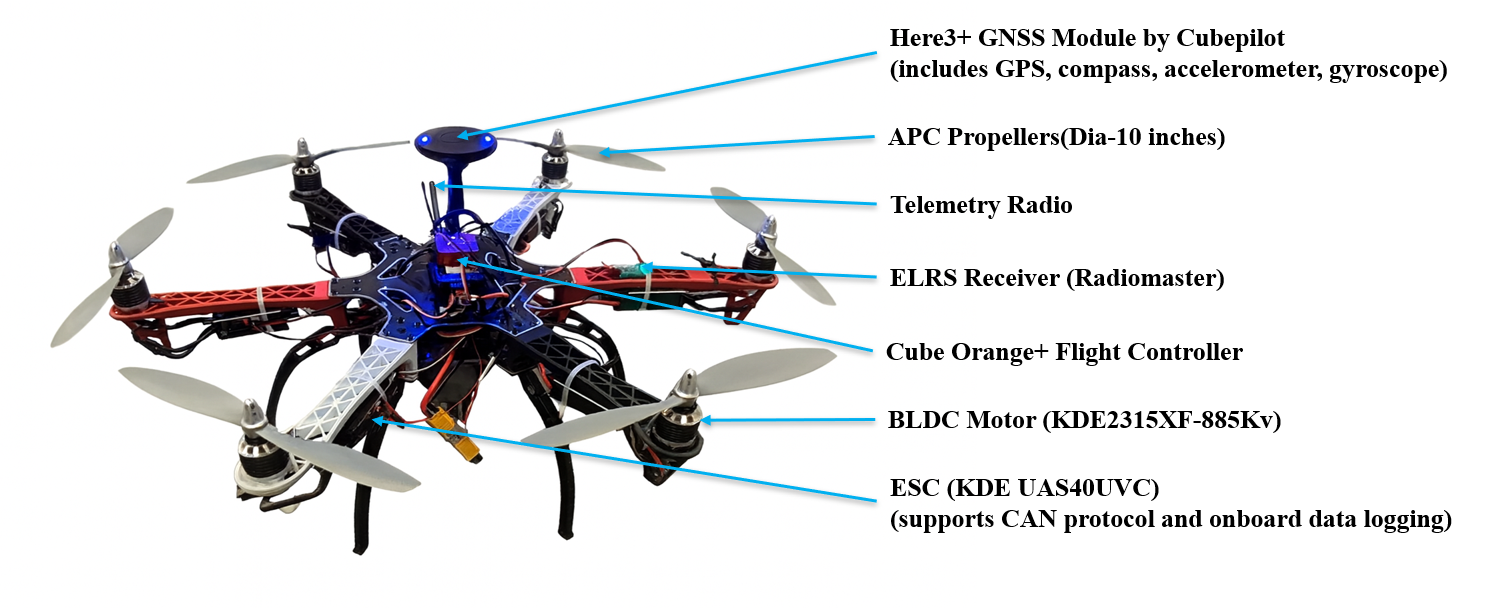}
    \captionof{figure}{Custom-built Hexacopter.}
    \label{fig:hexacopter}
\end{minipage}
As illustrated in Fig.~\ref{fig:data_acquisition}, the \textbf{Here3+ GNSS} module serves as the core navigation sensor, comprising four key sensing elements: GPS, magnetometer, gyroscope, and accelerometer. These sensors operate simultaneously to measure the vehicle’s spatial position, orientation, and motion rates. The raw data streams from the Here3+ sensors are processed through an \textbf{Extended Kalman Filter (EKF)} implemented within the Cube~Orange+ flight controller, which performs multi-sensor data fusion to estimate drift-corrected attitude, velocity, and position states.  
Following data fusion, the synchronized navigation and inertial data are stored through the Cube’s onboard logging system along with propulsion and power measurements. Propulsion telemetry is collected directly from the \textbf{KDE~UAS40UVC} electronic speed controllers (ESCs) via the CAN bus interface, providing real-time rotor speed (RPM), current, voltage, and temperature. Additionally, a \textbf{Cube~Power~Module} continuously measures the total current draw and battery voltage, enabling energy consumption analysis during flight operations.  All sensor and telemetry channels are sampled at different rate of frequency to produce high-resolution, time-synchronized multivariate datasets. These logs are later extracted using the Mission Planner interface and converted into \texttt{.CSV} format for preprocessing and model training. The integrated data acquisition system ensures accurate capture of vehicle kinematics, propulsion characteristics, and power signatures, forming the foundation for the experimental validation of the proposed fault detection framework.

\subsection{Flight Test Conditions and Data Collection}
\begin{figure}[!h]
    \centering
    \includegraphics[width=0.95\linewidth]{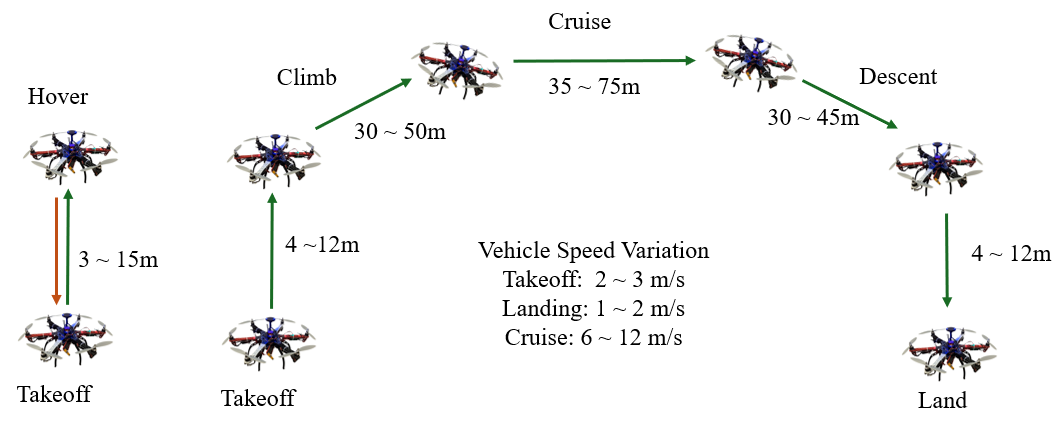}
    \caption{Flight test mission profiles: (left) short hover test for steady-state data collection; (right) full-trajectory mission including climb, cruise, descent, and landing phases.}
    \label{fig:flight_mission}
\end{figure}

Two types of flight missions were conducted to collect representative datasets for model training: (1) short-duration hover tests for propulsion and stability analysis, and (2) full-trajectory missions for dynamic flight characterization. These experiments were performed under calm wind conditions in an open field to minimize environmental disturbances. The overall mission profiles are illustrated in Fig.~\ref{fig:flight_mission}. Each experimental flight consists of an initial healthy operating segment followed by the introduction of a controlled actuator fault on a selected rotor channel. This structure enables direct comparison between nominal and faulty system behavior within the same flight.To ensure repeatability and statistical robustness, multiple flight trials were performed for each fault location and severity level. The resulting dataset contains both healthy and faulty operating conditions across all rotor channels, supporting supervised learning for fault classification and regression-based severity prediction.

\subsection{Propeller Fault Induction}
\begin{figure}[!h]
    \centering
    \includegraphics[width=0.95\linewidth]{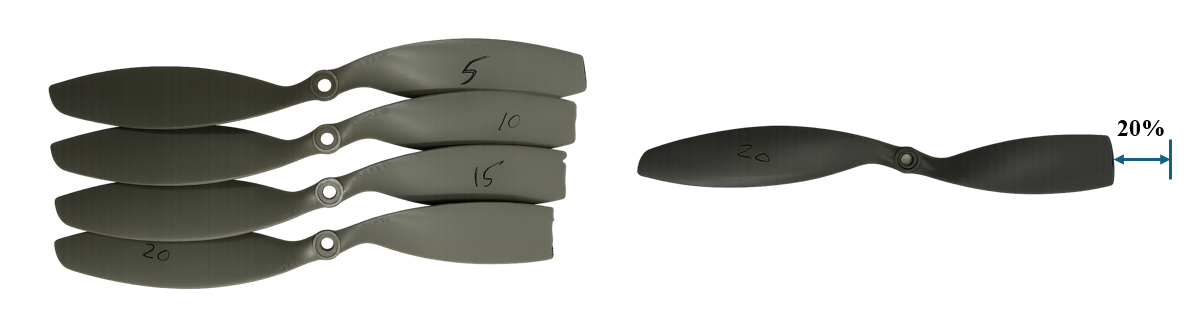}
    \caption{Propeller Fault Injection.}
    \label{fig:blade_fault_induction}
\end{figure}

To experimentally replicate realistic rotor faults and evaluate the proposed fault detection framework, controlled structural degradations were introduced on the propeller blades of the hexacopter. Since propeller damage or breakage directly affects thrust generation and aerodynamic balance, this approach allows systematic investigation of the dynamic and electrical responses of the vehicle under varying fault severities.As illustrated in Fig.~\ref{fig:blade_fault_induction}, the faults were simulated by cutting the blade tips of the APC~10×3.8~Slow~Flyer propellers at different lengths corresponding to \textbf{5\%}, \textbf{10\%}, \textbf{15\%}, and \textbf{20\%} of the total blade span. The removed portion was measured from the tip chord along the spanwise direction using a digital caliper to ensure repeatability across all blades. Each damaged propeller was labeled with its respective severity percentage and carefully balanced to minimize secondary vibration effects unrelated to the induced fault.During the experiments, one rotor at a time was replaced with a faulted propeller, while the remaining five rotors retained intact blades to maintain overall vehicle stability. This single–rotor–fault configuration enables the isolation of fault signatures associated with individual propulsion units. Tests were performed sequentially for all six rotors. The baseline (0\%) case was also included to represent nominal conditions for comparison.

\subsection{Data Processing}
\label{subsec:data_processing}

Raw flight data were recorded onboard the Cube~Orange+ flight controller including attitude (\texttt{ATT}), body angular rates (\texttt{RATE}), EKF navigation states (\texttt{XKF1}), and propulsion telemetry (\texttt{ESC}). A unified preprocessing pipeline was implemented in Python using the \texttt{pymavlink} library to extract, synchronize, and structure the data for learning-based fault diagnosis. From these logs, Euler angles \((\phi,\theta,\psi)\), angular rates \((p,q,r)\), inertial-frame positions \((P_N,P_E,P_D)\), velocities \((V_N,V_E,V_D)\), and six ESC RPM signals were extracted. All data streams were referenced to the onboard timestamp and normalized so that the first retained sample corresponded to \(t=0\). To remove preflight and post-landing segments, a bounded time window of interest was applied. Because the message groups were logged at different rates, temporal alignment was achieved via linear interpolation, using the \texttt{XKF1} navigation solution as the reference timeline. Inertial-frame velocities were transformed into body-frame components \((V_x,V_y,V_z)\) using a rotation matrix constructed from the measured roll, pitch, and yaw angles. After synchronization and transformation, all channels were merged into a single multivariate time series and samples with incomplete telemetry or zero RPM values were discarded. Each time step was represented by an 18-dimensional state vector comprising position, attitude, body-frame velocities, angular rates, and rotor speeds. The continuous time series was then segmented into fixed-length windows of \(N=100\) samples (10~s at 10~Hz) using a sliding-window approach, yielding an input tensor of dimension \(M \times N \times 18\). All features were normalized to \([0,1]\) using min--max scaling, and each window was labeled with a six-dimensional fault severity vector corresponding to the percentage of blade-tip breakage applied to each rotor. This processing pipeline preserves the coupled temporal evolution of vehicle dynamics and actuator behavior, enabling effective training of the CNN-based fault detection and severity prediction model.

\subsection{Model Training with Experimental Data}

The experimentally generated dataset introduced in the previous sections was employed to train and evaluate the proposed convolutional neural network (CNN) for rotor fault detection and severity estimation. The main goal of this stage was to assess the model’s predictive capability under real flight operating conditions. To ensure uniform representation of rotor positions and fault magnitudes, systematic tip-breakage experiments were performed for each motor across multiple damage severities. Specifically, the experimental dataset included nominal (0\%) operation as well as damaged cases with 5\%, 10\%, 15\%, and 20\% propeller tip breakage. Each motor--fault configuration was tested through multiple missions in both hover and full-trajectory flight regimes, resulting in a total of 24 distinct fault scenarios.

The CNN was trained from scratch to avoid bias and to allow learning directly from measured real-world flight data. Training utilized the \textbf{mean squared error (MSE)} loss function to reduce the discrepancy between predicted and ground-truth severity values for all six rotors. This training process enabled the network to capture the nonlinear mapping between observed vehicle states (attitude, body-frame velocities, angular rates, and rotor RPMs) and the corresponding propulsion health conditions reflected in the experimental measurements.

To maintain balanced inclusion of all fault types and severity levels, the dataset was randomly split into two subsets:
\begin{itemize}
    \item \textbf{Training set:} 80\% of the total samples, consisting of both nominal and faulted cases (0--20\%) for all six rotors.
    \item \textbf{Testing set:} 20\% of the total samples, consisting of both nominal and faulted cases (0--20\%) for all six rotors.
\end{itemize}

 \begin{figure}[!h]
    \centering
    \begin{minipage}{1\linewidth}
        \centering
        \includegraphics[width=\linewidth, height=5cm]{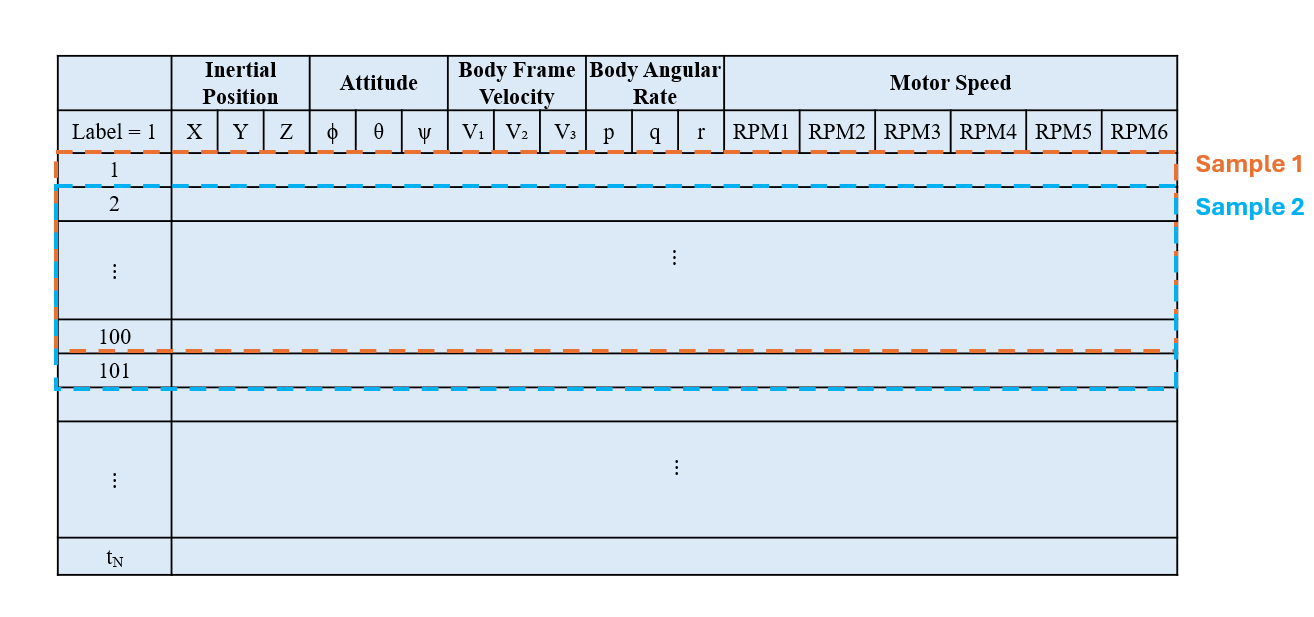}
        \caption{Sample input from generated dataset.}
        \label{fig:input_data_structure}
    \end{minipage}
\end{figure}

The trained CNN produces a six-dimensional output vector corresponding to the estimated fault severity (\(0\text{--}20\%\)) for each rotor. Model performance was quantified using the following metrics:
\begin{itemize}
    \item \textbf{Mean Absolute Error (MAE)} --- evaluates the per-rotor magnitude of prediction error.
    \item \textbf{Overall Accuracy (\%)} --- measures the percentage of correct severity predictions (across all six rotors) within a specified tolerance range.
\end{itemize}

Together, these evaluation measures provide a quantitative assessment of the CNN’s capability to generalize across different fault severities.

\begin{figure}[!h]
    \centering
    \begin{minipage}{1\linewidth}
        \centering
        \includegraphics[width= 0.9\linewidth]{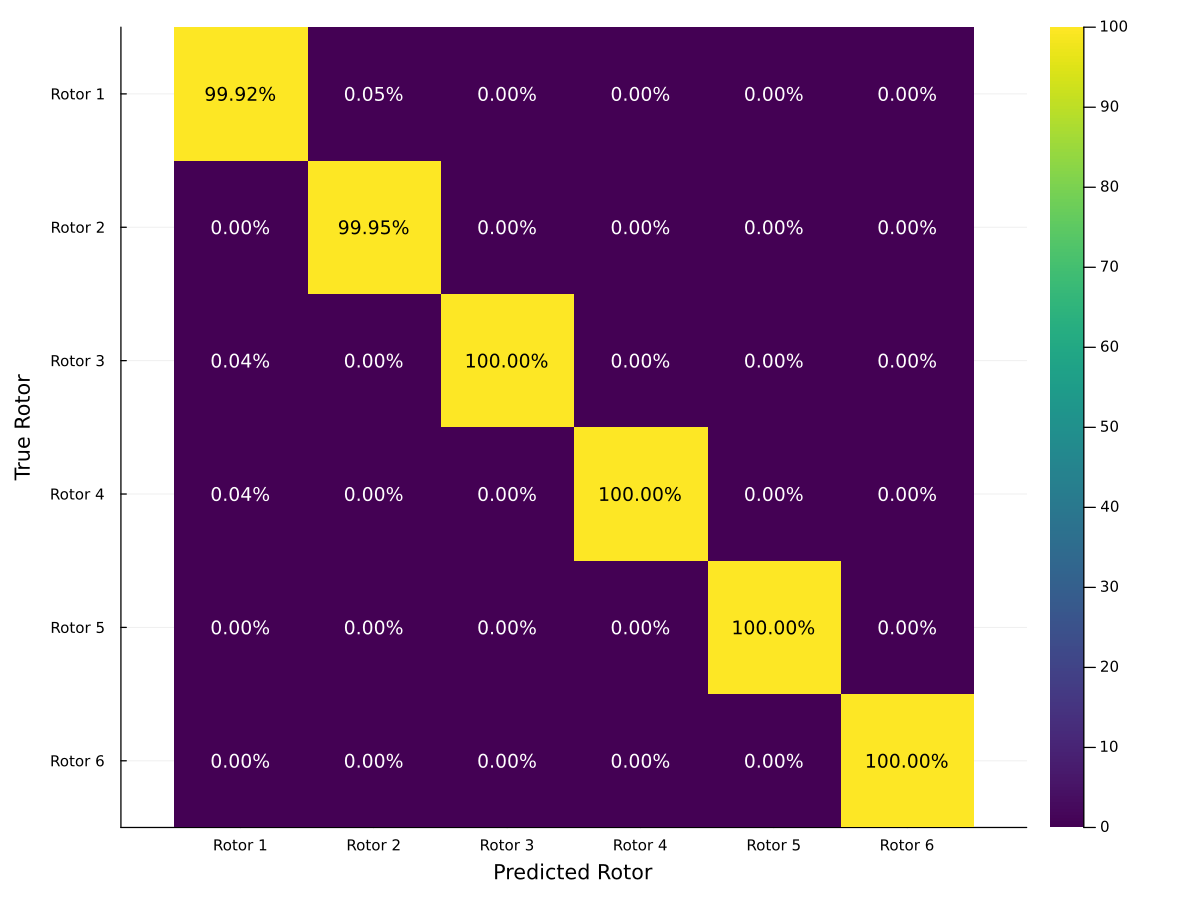}
        \caption{Confusion matrix of rotor prediction.}
        \label{fig:confusion_matrix_6_class}
    \end{minipage}
\end{figure}  

\section{Experimental Validation Results and Analysis}
 
The convolutional neural network (CNN) was trained on experimentally collected flight data for 300~epochs using a learning rate of $1\times10^{-5}$ and a batch size of 64. During training, the mean squared error (MSE) consistently decreased, and the average rotor-wise validation accuracy within a $\pm1\%$ damage tolerance of the ground-truth severity reached $94.46\%$.
\begin{figure*}[!h]
    \centering
    \begin{minipage}{1\linewidth}
        \centering
        \includegraphics[width=\linewidth, height=8cm]{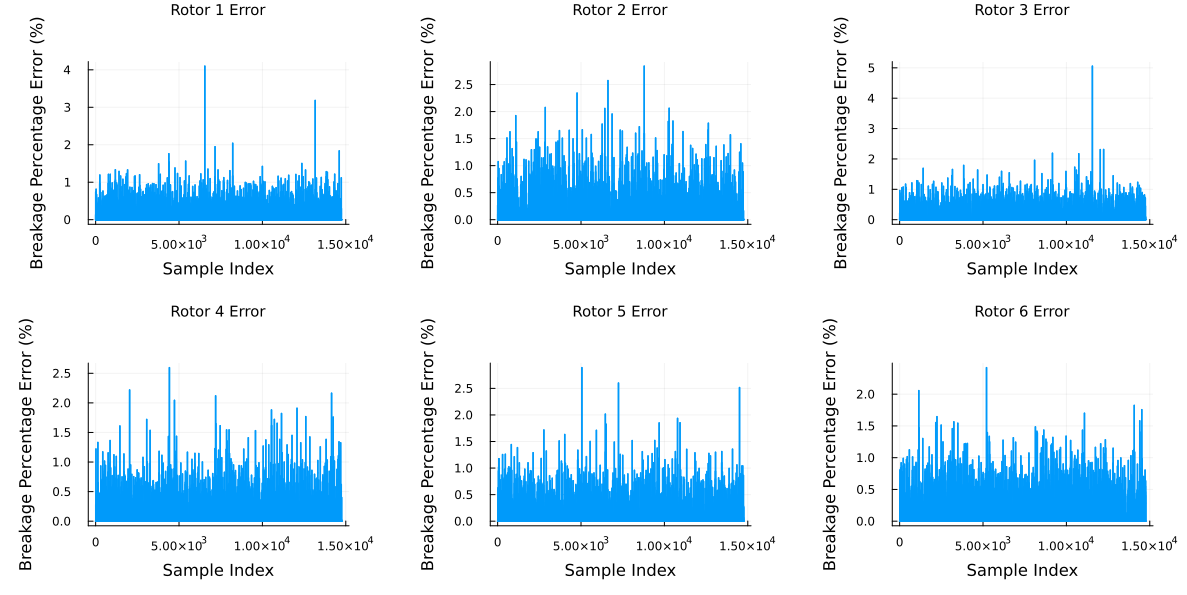}
        \caption{Rotor damage prediction errors for six rotors.}
        \label{fig:damage_errors}
    \end{minipage}
\end{figure*}
The prediction error for each rotor is defined as the absolute deviation between the estimated and true breakage levels, and is computed as:
\begin{equation}
\text{Damage error} = 100 \Big| y^{\text{pred}}_{i,r} - y^{\text{true}}_{i,r} \Big|.
\label{eq:prediction_error}
\end{equation}

where $i \in \{1,2,\dots,N_s\}$ and $r \in \{1,2,\dots,6\}$.

Across all six rotors, the trained model produced consistently strong testing performance. The rotor-wise test accuracies were 99.40\%, 99.15\%, 99.32\%, 99.46\%, 99.51\%, and 99.47\%, respectively. When requiring that all six rotors satisfy the $\pm1\%$ tolerance simultaneously, the resulting overall test accuracy was 96.39\%.

Rotor fault localization performance is further summarized using the six-class confusion matrix in Fig.~\ref{fig:confusion_matrix_6_class}, where rows denote the true faulty rotor and columns correspond to the predicted rotor index. The matrix is clearly dominated by diagonal entries, indicating that the CNN can accurately localize the damaged rotor over the full set of evaluated fault conditions. Off-diagonal entries are minimal, typically ranging from 0.00\% to 0.1\%, and there is no observable clustering of misclassification errors. In particular, rotors~3 to~6 demonstrate almost perfect classification, while rotors~1 and~2 also achieve excellent identification rates with only a small number of isolated incorrect predictions. The strong diagonal structure confirms that the spatio--temporal features learned by the network effectively capture the rotor-specific dynamic signatures associated with faults.

Figure~\ref{fig:damage_errors} presents the rotor-wise distribution of prediction errors across all damage percentages for each rotor. Most samples fall within 1\% of the actual damage level, showing that the model can reliably estimate subtle changes in rotor degradation during real flight experiments. A limited number of outliers exhibit slightly larger deviations, which can be attributed to effects such as sensor measurement noise, aerodynamic coupling/asymmetry, or uncertainties inherent to the flight-test environment. These remaining errors indicate that incorporating additional experimental data---especially covering a wider range of flight regimes and fault severities---could further improve generalization. Moreover, while the current results demonstrate strong predictive capability, the model has primarily been tested on fault scenarios that were partially represented during training. Therefore, further evaluation on completely unseen fault conditions, such as increased breakage levels (e.g., \(30\%\)), is required to more rigorously assess robustness and generalization performance.

\section{Conclusion and Future Work}
This paper introduces a data-driven framework for rotor fault detection, localization, and severity estimation in multirotor unmanned aerial systems (UAS), and demonstrates its implementation using a custom-developed multicopter test platform.  

Results from both simulation and experimental studies confirm that the proposed approach can successfully extract fault-related signatures from either simulated datasets or real-flight telemetry, enabling clear differentiation among healthy, degraded, and faulty propulsion conditions. In the experimental evaluation, minor discrepancies were observed in the predicted severity levels, particularly for small damage magnitudes. However, since the absolute fault severity is low in such cases, even small estimation offsets can appear relatively large when expressed as a percentage of the true fault magnitude. These deviations are likely caused by unavoidable experimental uncertainties such as outdoor disturbances, sensor noise, slight structural imbalance, and aerodynamic variations. While the trained model performs reliably within the fault severity range included in the training dataset, more extensive validation is still required using completely unseen fault scenarios, including higher breakage severities and fault combinations beyond those encountered during training.  

In general, this work contributes to a machine-learning-based framework capable of identifying rotor faults, determining their location, and estimating their severity, along with an experimental testbed that enables systematic evaluation of the proposed methodology. The combined framework and platform provide a solid foundation for autonomous propulsion health monitoring in UAV systems.

\subsection*{Recommendations for Future Work}

Several promising extensions can be explored to further improve the framework:

\begin{enumerate}
    \item \textbf{Further Validation Using Unseen Faulty Dataset:} 
     Future efforts should test the model on fully unseen fault cases, including greater breakage severities and fault patterns absent from the training data.

    \item \textbf{Model Environmental Disturbances:}  
    Incorporating wind estimation and gust-load effects would improve severity prediction accuracy during aggressive or highly dynamic flight conditions.

    \item \textbf{Adopt Hybrid Physics--ML Frameworks:}  
    Combining aerodynamic/physics-based modeling with machine learning could reduce the need for large datasets while enhancing interpretability and reliability.

    \item \textbf{Test Under Broader Flight Conditions:}  
    Additional validation under diverse mission conditions---including changes in payload, speed, weather, and operational profiles---would improve robustness and field readiness.

    \item \textbf{Include Diverse Fault Types:}  
    The framework can be extended to account for broader fault modes, such as bent propellers, progressive motor degradation, and sensor-related failures, improving overall applicability and resilience.

\end{enumerate}

Collectively, these directions offer a pathway toward more robust and fault-tolerant autonomous aerial systems with improved capability for early detection, prediction, and mitigation of propulsion system failures.

\bibliography{fdi_references}

\end{document}